\documentclass{article}
\usepackage{ijcai23}

\usepackage{times}
\usepackage{soul}
\usepackage{url}
\usepackage[hidelinks]{hyperref}
\usepackage[utf8]{inputenc}
\usepackage[small]{caption}
\usepackage{graphicx}
\usepackage{amsmath}
\usepackage{amsthm}
\usepackage{booktabs}
\usepackage{amsfonts}
\usepackage{algorithm}
\usepackage{algorithmic}
\usepackage[table]{xcolor}
\usepackage{todonotes}
\usepackage{pifont}
\usepackage{multirow}
\usepackage[switch]{lineno}
\usepackage{blindtext}

\title{Towards Explainable and Safe Conversational Agents for Mental Health:\\ A Survey}

\begin{document}
\author{
Surjodeep Sarkar$^1$\and
Manas Gaur$^1$\and
Lujie Karen Chen$^1$\and \\
Muskan Garg$^2$\and 
Biplav Srivastava$^3$\and
Bhaktee Dongaonkar$^4$
\affiliations
$^1$UMBC, MD, USA ;
$^2$Mayo Clinic, MN, USA; 
$^3$AI Institute, USC, SC, USA:
$^4$IIIT Hyderabad, India
\emails
\{ssarkar1, manas, lujiec\}@umbc.edu,
biplav.s@sc.edu, \\ garg.muskan@mayo.edu, bhaktee.dongaonkar@iiit.ac.in
}

\maketitle



\begin{abstract}

Virtual Mental Health Assistants (VMHA) are seeing continual advancements to support the overburdened global healthcare system that gets 60 million primary care visits, and 6 million Emergency Room (ER) visits annually. These systems are built by clinical psychologists, psychiatrists, and Artificial Intelligence (AI) researchers for Cognitive Behavioral Therapy (CBT). At present, the role of VMHAs is to provide emotional support through information, focusing less on developing a reflective conversation with the patient. A more \textit{comprehensive, safe} and \textit{explainable} approach is required to build \textit{responsible} VMHAs to ask follow-up questions or provide a well-informed response. This survey offers a systematic critical review of the existing conversational agents in mental health, followed by new insights into the improvements of VMHAs with contextual knowledge, datasets, and their emerging role in clinical decision support. We also provide new directions toward enriching the user experience of VMHAs with explainability, safety, and wholesome trustworthiness. Finally, we provide evaluation metrics and practical considerations for VMHAs beyond the current literature to build trust between VMHAs and patients in active communications.

\end{abstract}

\section{Introduction}
Mental illness is highly prevalent nowadays, constituting a major cause of distress in people’s lives with an impact on society’s health and well-being, thereby projecting serious challenges for mental health professionals (MHPs)~\cite{zhang2022natural}. According to the National Survey on Drug Use and Health, nearly one in five U.S. adults live with a mental illness (52.9 million in 2020)~\cite{SAMSHA}. Reports released in August 2021\footnote{https://www.theguardian.com/society/2021/aug/29/strain-on-mental-health-care-leaves-8m-people-without-help-say-nhs-leaders} indicate that \textit{1.6 million people} in England were on waiting lists to seek professional help with mental health care. Such an overwhelming rise in the number of patients as compared to MHPs necessitated the 
use of (i) public health forums~(e.g., dialogue4health), (ii) online communities~(e.g., r/depression subreddit on Reddit), (iii) Talklife, and (iv) Virtual Mental Health Assistants (VMHAs), for informative healthcare. The anonymous functioning of (i), (ii), (iii) removed the psychological stigma in patients, which even refrained them from seeing an MHP \cite{hyman2008self}. 

In addition, the unavailability of interpersonal interactions from other pure information agents resulted in the need to develop Virtual Mental Health Assistants (VMHAs).
\begin{figure}[t]
    \centering
\includegraphics[width=0.45\textwidth]{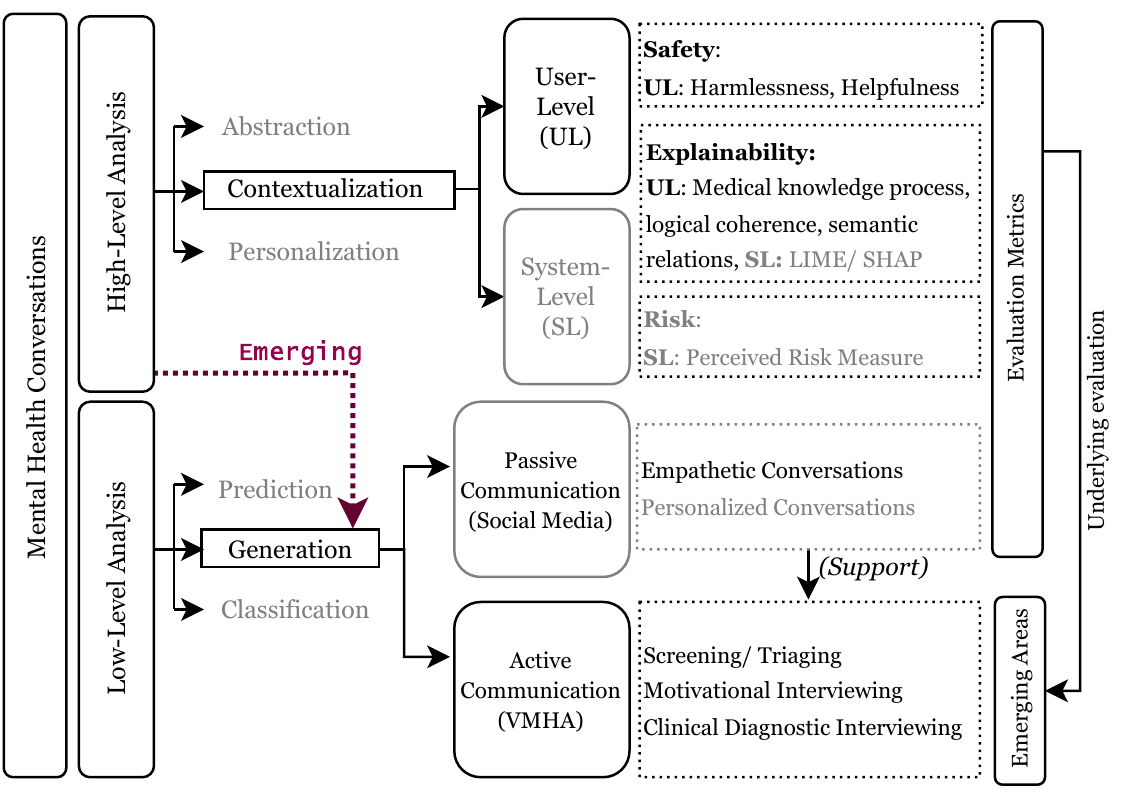}
    \caption{Taxonomy of Mental Health Conversations: While connecting dots in our investigation from NLP-centered low-level analysis (lexical, morphological, syntactic, semantic) over Mental health conversations to the higher-level analysis (discourses, pragmatics), we determine the evaluation metrics to support VMHAs for better user-level experience in terms of safety and explainability. We further support the emerging areas with AI model development and evaluation in passive conversations. The categories in black color defines the scope of our survey from the view point of user-level explainability and safety; dotted red colour highlights the emerging scope of question/response generation in mental health conversations between VHMAs and patients. }
    \label{fig:1}
\end{figure}

\begin{figure*}[!ht]
\centering
\includegraphics[width=\textwidth]{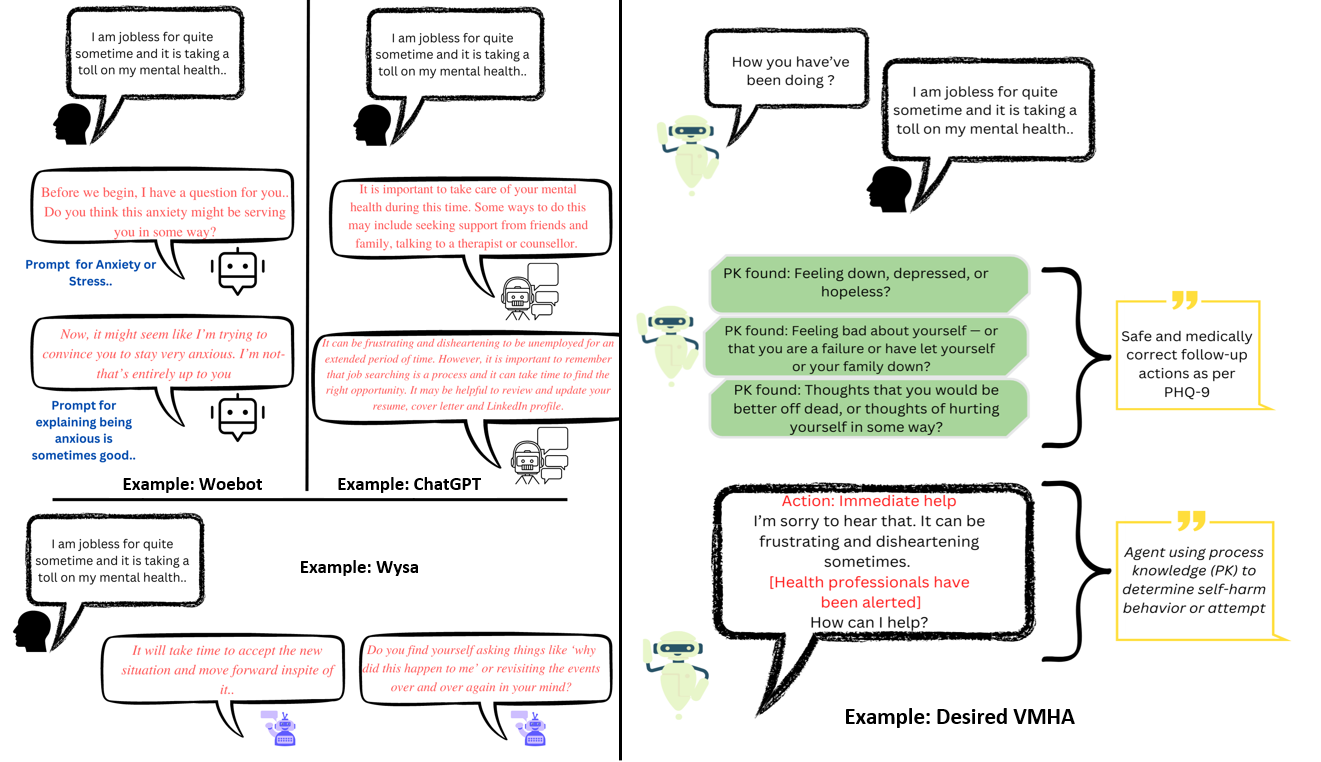}
\caption{(Left) The outcome from existing VMHAs (e.g., WoeBot, Wysa) and ChatGPT (general purpose chatbot). (Right) Illustration of a knowledge-driven conversational agent in mental health (desired VMHA). The use of questions in PHQ-9 to induce conceptual flow in mental health conversational agents. With clinical knowledge, the agent can detect the user's mental disturbance and alert MHPs accordingly.}
\label{fig:conv}
\end{figure*}

\noindent \textbf{VMHAs}: VMHAs are artificial intelligence (AI)-based agents designed to provide emotional support through structured conversational sequences targeted to screen patients for mental health conditions and alert mental health professionals (MHPs) through \textit{informed triaging}\footnote{\url{https://code4health.org/chat-bot/}}. Despite the proliferation of research at the intersection of clinical psychology, artificial intelligence (AI), and Natural Language Understanding (NLU), VMHAs missed an opportunity to serve as life-saving contextualized, personalized, and reliable decision support during COVID-19 under the \textit{apollo} moment~\cite{apollo-chatbots,czeisler2020mental}. VMHAs' ability to function as simple information agents (e.g., suggest meditation, relaxation exercises, or give positive affirmations) \textit{did not} bridge the gap between \textit{monitoring the health condition} and \textit{necessitating an MHP visit} for the patient.

To the best of our knowledge, this is the first critical evaluation that examines contextualization and question/response generation for VMHAs from the viewpoint of user-level explainability and trust (see Figure \ref{fig:1}). \textit{This survey facilitates the clinical psychologists, psychiatrists, and AI practitioners of VMHAs to support people at risk of chronic mental disease.}

\noindent \textbf{User-level Explainability}: The sensitive nature of VMHAs raises \textit{safety} as a major concern of conversational systems, resulting in a negative outcome. For instance, figure \ref{fig:conv} presents a real-world query from a user, which was common during the times of the COVID-19 recession.
In response to the query- Woebot, Wysa and ChatGPT initiated a responsive conversation without focusing on the context (e.g., connecting mental health with its symptoms). We found assumptive questions (e.g., anxiety) and responses from Wysa, Woebot and ChatGPT with no association to clinical reference or clinical support. On the other hand, the desired VMHA (a) should capture the relationship between the user query and expert questionnaires and (b) tailor the response to reflect on the user's concerns (e.g., \textit{frustrating} and \textit{disheartening}) about the \textit{long-term unemployment}, which is linked to \textit{mental health} and \textit{user immediate help}. 


\noindent \textbf{Resources to support VMHA}: Prior research demonstrate extensive body of efforts in developing \textit{mental health datasets} using social media to identify mental health conditions~\cite{uban2021emotion}. These 
datasets represent real-world conversations and are annotated by experts leveraging clinically-grounded knowledge (e.g., MedChatbot~\cite{kazi2012medchatbot}) or guidelines (e.g., PHQ-9). Augmenting such datasets with VMHAs can improve the quality of conversations with the user. Semantic enhancements with clinical knowledge and associated guidelines, if remain under-explored, may miss the hidden mental states in a given narrative which is an essential component of question generation.


\noindent \textbf{Trustworthiness}: By definition, \textit{Trust} is a multi-faceted quality that is studied in the context of humans in humanities and now increasingly gaining importance in AI as systems and humans collaborate closely. Growing concern about (misplaced) \textit{trust} on \textit{VMHA} for \textit{Social Media} (tackling mental health) hampers the adoption of AI techniques during emergency situations like COVID-19~\cite{apollo-chatbots}. 
A recent surge in the use of ChatGPT, in particular for mental health, is emergent for providing crucial personalized advice without clinical explanation, which might hurt user's \textit{safety}, and thus, \textit{trust} \footnote{\url{https://tinyurl.com/4sr2hw9b}}. In~\cite{Varshney2022}, the author identifies the support for human interaction and explainable alignment with human values as important for trust in AI systems.  

To holistically contribute towards \textit{trustworthy} behavior in a conversational system in mental health, there is a need to critically examine \textit{user-level explainability}, \textit{safety}, the use of clinical knowledge for contextualization, along with testing. 

\noindent \textbf{Our Contributions}: This survey spans 
5 major research dimensions: (i) What are explainability and safety in VMHAs? (ii) What are the current capabilities and limitations of VMHA?, (iii) What is the current state of AI and the hurdles in supporting VMHAs? (iv) What functionalities can be imagined in VMHA for which patients seek alternative solutions? and (v) What changes in evaluation is required with respect to explainability, safety, and trust? Figure \ref{fig:1} illustrates the survey coverage, exemplified in Figure \ref{fig:conv}.


\section{Scope of Survey} 
In this section, we explore the state of research in explainability and safety in conversational systems to ensure trust~\cite{trust-explainablemetric-survey}.

\subsection{Explanation}

\begin{figure}[!t]
    \centering
\includegraphics[width=0.50\textwidth]{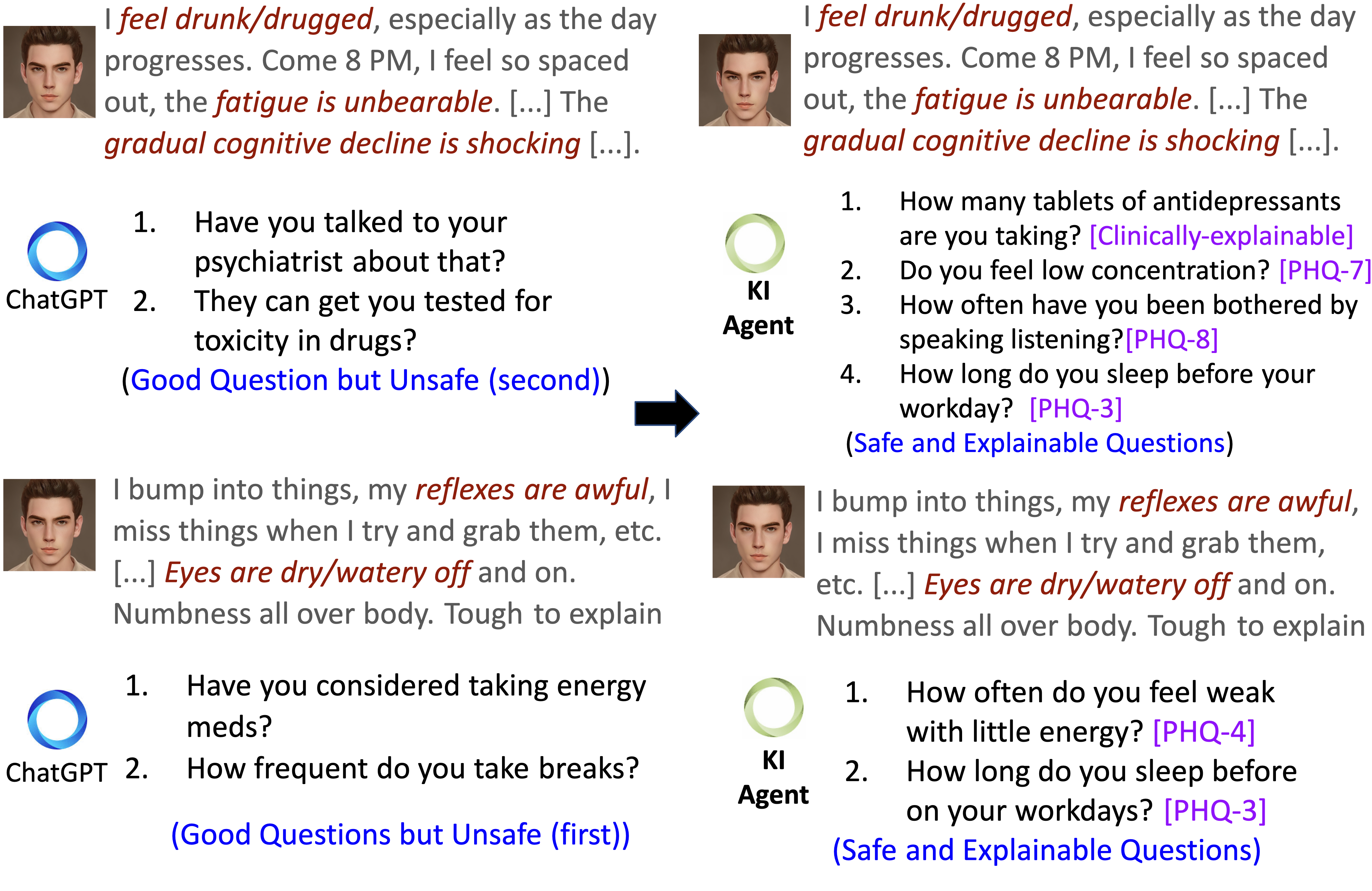}
    \caption{A conversational scenario in which a user asks a query with multiple symptoms. Left is a set of generated questions obtained by repetitive prompting ChatGPT. Right is a generation from ALLEVIATE, a knowledge-infused (KI) conversational agent with access to PHQ-9 and clinical knowledge from Mayo Clinic.}
    \label{fig:safe}
\end{figure}

Conversations in AI happen through large and complex language models (e.g., GPT-3, ChatGPT), which are established as state-of-the-art models for developing intelligent agents to chat with the users by generating human-like questions or responses. The reasons behind the output generated by the Large Language Models (LLM) are unclear and hard to interpret, also known as the ``\textit{black box}'' effect. The consequences of the black box effect are more concerning than their utility, particularly in mental health. Figure \ref{fig:safe} presents a scenario where ChatGPT advises the user about \textit{toxicity in drugs}, which may have a negative consequence. 
To this end, \cite{bommasani2021opportunities} reports hallucination and harmful question generations as unexpected behaviors shown by such black box models. The study characterizes \textit{hallucination} as a generated content that \textit{deviates} significantly from the subject matter or is unreasonable. Recently, Replika
, a VMHA, augmented with a GPT-3, provides meditative suggestions to a user expressing self-harm tendencies\footnote{\url{https://ineqe.com/2022/01/20/replika-ai-friend/}}. 
The analysis above supports the critical need for a comprehensive and explainable approach toward the decision-making of VMHAs. According to~\cite{weick1995sensemaking}, the explanations are human-centered sentences that signify the reason or justification behind an action and are comprehensible to a human expert. There are many types of explanations \cite{explainable-ai-survey} and surveys of deployed systems \cite{explanability-users-survey} has revealed that most are targeted towards model developers and not the end-users. The users interacting with the VMHAs may need more systematic information than just the decision-making. Thus, this survey is more focused towards ``\textit{User-level Explainability}''.

\noindent \textbf{User-level Explainability (UsEx)} \emph{is defined as the capability of an AI methodology to provide a post-hoc explanation upon the need of a user and in the form of traceable links to real-world entities and definitions~\cite{gaur2022knowledge}.}

Figure \ref{fig:safe} illustrates the UsEx wherein the generated follow-up questions from a safe and user-level explainable agent establish semantic connections with clinical guidelines (e.g., PHQ-9). Though UsEx sees promise over foundational general-purpose NLP tasks, its applicability in the mental health context is yet to be examined~\cite{gunaratna2022explainable}.   

\subsection{Safety}

VMHAs are required to be predominantly safe while at the same time being explainable to prevent undesirable behaviors. One such method is aligning the functioning of VMHA to MHP-defined specifications \cite{koulouri2022chatbots}. Such specifications allow VMHAs absolve the control of generating fabricated content and render it unsafe.

~\cite{dinan2021anticipating} identifies three major effects on safety in general-purpose conversational systems: (a) Generating Offensive Content, also known as the \textit{Instigator (Tay) Effect}. It describes the tendencies of a conversational agent to display behaviors like the Microsoft Tay chatbot, which went racial after learning from the internet. (b) \textit{YEA-SAYER (ELIZA)} effect is defined as the response from a conversational agent to an offensive input from the user. People have been proven to be particularly forthcoming about their mental health problems in interactions with conversational agents, which may increase the danger of ``\textit{agreeing with user utterances implying self-harm}''. (c) \textit{Imposter} effect applies to VMHAs that tend to respond \textit{inappropriately} in sensitive scenarios.
To overcome the imposter effect, Deepmind designed \textit{Sparrow}, a conversational agent. It responsibly leverages the live google search to talk with users~\cite{gupta-2022}. The agent generates answers by following the \textit{23 rules} determined by researchers, such as \textit{not offering financial advice}, \textit{making threatening statements}, or \textit{claiming to be a person}~\cite{heikkilä-2022}.

In the context of mental health, such rules can be replaced by clinical specifications to validate the functioning of AI model within the \textit{safe limits}. Source for such specifications are: Systematized Nomenclature of Medicine-Clinical Terms (SNOMED-CT)~\cite{donnelly2006snomed}, International Classification of Diseases (ICD-10)~\cite{quan2005coding}, Diagnostic Statistical Manual for Mental Health Disorder (DSM-5)~\cite{regier2013dsm}, Structured Clinical Interviews for DSM-5 (SCID)~\cite{first2014structured}, and clinical questionnaire-guided lexicons.~\cite{hennemann2022diagnostic} performs a comparative study on psychotherapy of outpatients in mental health where an AI model within VMHA aligns to clinical guidelines for easy understanding of domain experts through UsEx.



\section{Knowledge Infused (KI) Learning for Mental Health Conversations}
Machine-readable knowledge can be categorized into five forms: (a) lexical and linguistic, (b) general-purpose (e.g., Wikipedia, Wikidata), (c) commonsense (e.g., ConceptNet), (d) domain-specific (Unified Medical Language System), and (e) procedural or process-oriented (\textbf{PK})~\cite{sheth2021duality}. Knowledge-infused Learning (KIL), a paradigm within AI, defines a set of methodologies incorporating these broader forms of knowledge to address the limitations of current black-box AI. In addition, KiL benefits from data and knowledge to enable safe and explainable operations in mental health~\cite{gaur2022knowledge}. We categorize the KIL-driven efforts at the intersection of conversational AI and mental health into two categories: 

\begin{itemize}
    \item \textbf{Knowledge Graph-guided Conversations:} Question answering using knowledge graph (KG) is seeing tremendous interest from AI and NLU community through various technological improvements in query understanding, query rewriting, knowledge retrieval, question generation, response shaping, and others. The methods proposed can improve the high-level functionalities of VMHA. For instance, ~\cite{welivita2022heal}'s HEAL KG can generate a better empathetic response by capturing empathy, expectations, affect, stressors, and feedback types from distress conversations. With HEAL, the model picks an appropriate phrase in the user's query to tailor its response. EmoKG is another KG that connects BioPortal, SNOMEDCT, RxNORM, MedDRA, and emotion ontologies to have a conversation with a user to boost their mental health with food recommendation~\cite{gyrard2022interdisciplinary}. Likewise,~\cite{cao2020building} created suicide KG to train conversational agents that can sense whether the user interacting has suicidal indication (e.g., relationship issues, family issues) or suicide risk tendencies before yielding a response or asking follow-up questions.~\cite{sheth2019extending} explained the evolution of KG in VMHA during a conversation for adaptive communications. Augmentation of KG demands improvement in metrics to examine the safety and user-level explainability through proxy measures such as logical coherence, semantic relations, and others (covered in section \ref{eval} and~\cite{gaur2022iseeq}). 
     \item \textbf{Lexicon or Process-guided Conversations:} Lexicons in mental health were created to resolve ambiguities in human language. For instance, the following two sentences: ``I am feeling on the edge.'' and ``I am feeling anxious,'' are similar, provided there is a lexicon with ``Anxiety'' as a category and ``feeling on the edge'' as its concept.~\cite{yazdavar2017semi} created a PHQ-9 lexicon to study realistic mental health conversations on social media clinically.~\cite{roy2022proknow} leveraged PHQ-9 and SNOMED-CT lexicons to train a question-generating agent for paraphrasing questions in  PHQ-9 to introduce \textit{Diversity in Generation} \textbf{(DiG)}~\cite{limsopatham2016normalising}. With DiG, a VMHA can paraphrase its question to acquire a meaningful response from a user while still keeping engagement. \textit{Clinical specifications\footnote{also called clinical guidelines and clinical process knowledge}}(PK) include questionnaires such as PHQ-9 (depression), Columbia Suicide Severity Rating Scale (C-SSRS; suicide), Generalized Anxiety Disorder (GAD-7)~\cite{gaur2022knowledge}. It provides a sequence of questions clinicians follow to interview patients. Such questions are safe and medically validated. ~\cite{noble2022developing} developed MIRA, a VMHA with knowledge of clinical specification to meaningfully respond to queries on mental health issues and interpersonal needs during COVID-19.~\cite{miner2016conversational} leverage Relational Frame Theory (RFT), a procedural knowledge in clinical psychology to capture events between conversations and labels as positive and negative.~\cite{chung2021chatbot} develops KakaoTalk, a chatbot with prenatal and postnatal care knowledge database of Korean clinical assessment questionnaires and responses that enable the VMHA to carry out thoughtful and contextual conversations with users. 
\end{itemize}

Using KGs through mechanisms of KIL can propel context understanding in VMHA for a safe and explainable conversation. 
Datasets or VMHAs which use mental health-related knowledge (\textbf{MK}) as either KG or lexical are marked as \ding{51} in Tables \ref{tab:datasets} and \ref{tab:my_label}.

\begin{table*}[!ht]
    \centering
    \begin{tabular}{lccc|ccc|cccc}
    \toprule[1.5pt]
        \multicolumn{2}{c}{Datasets}  &  Safety & UsEx & \multicolumn{2}{c}{KI} & DiG & \multicolumn{4}{c}{FAIR Principle} \\
        & & & & PK & MK & & F & A & I & R\\ \midrule
        {~\cite{counselchat}} & CounselChat & \ding{51} &  \ding{55} & \ding{55} & \ding{55}& \ding{55} & \ding{51} & \ding{51} & \ding{55}& $\dagger$\\
        {~\cite{huang2015language}} & CC & \ding{55} & \ding{51} & \ding{55} & \ding{51}  & \ding{55} & \ding{51} & \ding{51} &\ding{55} & $\dagger$ \\
        {~\cite{althoff2016large}} & SNAP Counseling & \ding{51} & \ding{55} & \ding{55} &  \ding{55}& \ding{51} & \ding{55} & \ding{55} & \ding{55}& \ding{55}\\
        {~\cite{rashkin2019towards}} & Empathetic Dialogues & \ding{51} & \ding{55} & \ding{55} & \ding{55} & \ding{51} & \ding{51} & \ding{51} & \ding{51}& \ding{51} \\
        {~\cite{demasi2019towards}} & Roleplay & \ding{51} & \ding{51} &\ding{51} &\ding{55} & \ding{51} & \ding{51} & \ding{51} & \ding{55}& \ding{51}\\
        {~\cite{liang2021evaluation}} & CC-44 &  \ding{55} & \ding{55} & \ding{55} & \ding{55} & \ding{55} & \ding{51} & $\dagger$ & \ding{55}& $\dagger$ \\
        {~\cite{gupta2022learning}} & PRIMATE & \ding{51} & \ding{51} & \ding{51} & \ding{55} & \ding{55} & \ding{51} & \ding{51} & \ding{51}& \ding{51} \\
        {~\cite{roy2022proknow}} & ProKnow-data & \ding{51} & \ding{51} & \ding{51} & \ding{51} & \ding{51} & \ding{51} & \ding{51} & \ding{51}&  \ding{51}\\
        {~\cite{welivita2022curating}} & MITI & \ding{51} & \ding{51} & \ding{55} & \ding{55} & \ding{55} & \ding{51} & \ding{51} & \ding{51} & \ding{51}
        \\ \bottomrule[1.5pt] 
    \end{tabular}
    \caption{Lists of conversational datasets created with support from MHPs, crisis counselors, nurse practitioners, or trained annotators. We have not included datasets created using crowdsource workers without proper annotation guidelines. KI: Knowledge infusion; PK: Process Knowledge; MK: Medical Knowledge; DiG: Diversity in Generation; UsEx: User-level Explainability. Here, The \textit{FAIR principles} stands for F: Findability, A: Accessibility, I: Interoperability, and R: Reusability. $\dagger$: partial fulfillment of the corresponding principle.}
    \label{tab:datasets}
\end{table*}

\section{Safe and Explainable Language Models in Mental Health}
Language models (e.g., Blenderbot, DialoGPT) and in-use conversational agents (e.g., Xiaoice, Tay, Siri) were questioned in the context of safety during the \textit{first workshop on safety in conversational AI}. ~70\% participants in the workshop were unsure of whether present-day conversational systems or language models within them are capable of safe generation. Following it,~\cite{xu2020recipes} introduced \textit{Bot-Adversarial Dialogue} and \textit{Bot Baked In} methods to introduce \textit{safety} in conversational systems. The study was performed on \textit{Blenderbot}, which had mixed opinions on safety, and \textit{DialoGPT}, to enable AI models to detect unsafe/safe utterances, avoid sensitive topics, and provide responses that are gender-neutral. The study utilizes knowledge from Wikipedia (for offensive words) and knowledge-powered methods to train conversational agents~\cite{dinan2018wizard}. Alternatively, safety in conversational systems can be introduced through clinical guidelines.
~\cite{roy2022proknow} develop safety lexicons from PHQ-9 and GAD-7, for safe and explainable functioning of language models. The study showed an 85\% improvement in safety across Sequence to Sequence and Attention-based language models. In addition, explainability saw an uptake of 23\% across the same language models. Similar results were observed when PHQ-9 was used in explainable training of language models~\cite{zirikly2022explaining}. 
VMHA can align with clinical guidelines through reinforcement learning. For example, the \textit{policy gradient-based learning} can assist conversational systems in accounting for safe generation, either through specialized datasets on response rewriting~\cite{sharma2021towards} or tree-based rewards guided by process knowledge in mental health~\cite{roy2022process}.  

Though there is an initiative to attain safety in conversations from AI-powered agents, an effort is needed to achieve UsEx. In mental health, the indicators of signs and symptoms, causes, disorders, medications, and other comorbid conditions possess probabilistic relationships with one another. Hence, the augmentation of the knowledge base or infusion of knowledge to improve AI's decision-making is crucial to human understandability~\cite{joyce2023explainable}.



\section{Virtual Mental Health Assistants}
Despite the positive potentials of the language models, our observations indicate the in-capabilities of VMHAs to comprehend the behavioral and emotional instability, self-harm tendencies, and user's latent psychological mindset. VMHAs (e.g., as exemplified in Figure \ref{fig:safe} and \ref{fig:conv}) generate incoherent and unsafe responses when a user tries to seek a response for clinically relevant questions. In this section, we outline the capabilities of well-established VMHAs and inspect limitations in the context of UsEx and safety following taxonomy in figure \ref{fig:1}. 

\begin{itemize}
    \item \textsc{WoeBot} is introduced as a part of the growing industry of digital mental health space as an ``\textit{Automated Coach}'' that can deliver a coach-like or sponsor-like experience without the human intervention to facilitate the ``\textit{good thinking hygiene}''~\footnote{\url{https://woebothealth.com/why-we-need-mental-health-chatbots/}}. \textsc{WoeBot} deploys featuring lessons (via texts and ``stories''), interactive exercises, and videos that were tuned around Cognitive Behavioral Therapy (CBT)~\cite{fitzpatrick2017delivering}. 
    \item \textsc{Wysa}, a mental health application, uses CBT conversational agent to have empathetic/ therapeutic conversations and activities thereby helping its users with several mental health problems~\cite{inkster2018empathy}. 
    Based on a series of question-answering mechanisms, Wysa suggests a set of relaxing activities for elevating mental well-being.

\end{itemize}




\begin{table*}[!ht]
\footnotesize
    \centering
    \begin{tabular}{lccccc|cc|l}
    \toprule[1.5pt]
        \multicolumn{2}{c}{VMHA} & Objective & \multicolumn{2}{c}{KI} & DiG & Safety & UsEx & QM  \\
         &  & &  PK & MK & & & \\ \midrule
         {~\cite{ginger}} & Ginger & Behavioral Health Coaching & \ding{55}  & \ding{51} & \ding{55} & \ding{55} & \ding{55} & H \\
         {~\cite{CompanionMX}} & CompanionMX & PTSD & \ding{55}  & \ding{55} & \ding{55} & \ding{55} & \ding{55} & H \\
         {~\cite{quartet}} & Quartet & Therapy \& Counseling & \ding{55}  & \ding{55} & \ding{55} & \ding{55} & \ding{55} & H\\ 
         {~\cite{fitzpatrick2017delivering}} & Woebot & CBT & \ding{51}  & \ding{51} & \ding{55} & \ding{55} & \ding{55} & A\\
         {~\cite{limbic}} & Limbic & CBT & \ding{55}  & \ding{55} & \ding{55} & \ding{51} & \ding{55} & H\\
         {~\cite{inkster2018empathy}} & Wysa & CBT & \ding{55}  & \ding{55} & \ding{55} & \ding{55} & \ding{55} & A\\
         {~\cite{fulmer2018using}} & Tess & Anxiety \& Depression & \ding{55} & \ding{55} & \ding{55} & \ding{55} & \ding{55}& - \\
         {~\cite{ghandeharioun2019emma}} & EMMA & CBT & \ding{55} & \ding{55} & \ding{55} & \ding{55} & \ding{55} & H \\
         {~\cite{denecke2020mental}} & SERMO & CBT & \ding{55} & \ding{55} & \ding{55} & \ding{55} & \ding{55} & H\\
        {~\cite{possati2022psychoanalyzing}} & Replika & Empathetic \& Supportive & \ding{55} & \ding{55} & \ding{55} & \ding{55} & \ding{55} & A \\ 
         {~\cite{roy2023alleviate}} & ALLEVIATE & Depression & \ding{51}  & \ding{51} & \ding{51} & \ding{51} & \ding{55} & H
        \\ \midrule
         Our Survey Paper & Desired System & Screening, Triaging, \& MI & \ding{51}  & \ding{51} & \ding{51} & \ding{51} & \ding{51} & H,A,T \\ \bottomrule[1.5pt]
    \end{tabular}
    \caption{Prominent and in-use VMHAs with different objectives for supporting patients with mental disturbance. We performed a high-level analysis of all the VMHAs based on publicly-available user reviews on forums (e.g., WebMD, AskaPatient, MedicineNet), and Reddit. For Woebot, Wysa, and Alleviate, a survey of 40 participants was carried out at Prisma Health. Here we define QM: Qualitative Metrics as H: Harmlessness, A: Adherence, T: Transparency.}
    \label{tab:my_label}
\end{table*}

With the historical evolution of VMHAs (see Table~\ref{tab:my_label}) from behavioral health coaching~\cite{ginger} to KG-based intellectual VMHAs such as ALLEVIATE~\cite{roy2023alleviate}, we examine the possibilities of new research directions to facilitate the expression of empathy in passive communications~\cite{sharma2023human}. The existing studies suggest the risk of oversimplification of mental conditions and therapeutic approaches without considering latent or external contextual knowledge~\cite{cirillo2020sex}. Thinking beyond the low-level analysis of classification and prediction, the high-level analysis of VMHAs would enrich the User-Level (UL) experience and informedness of MHPs~\cite{roy2023alleviate}.

Limiting our discovery to context-based high-level analysis, the System-Level (SL) observations for \textsc{WoeBot and Wysa} suggest the UL tracking of human behavior, such as gratitude/ mindfulness and frequent mood changes (an emotional spectrum) during the day. Contributions toward this endeavor have emerged through exclusive studies with \textit{trustworthiness} of WoeBot and Wysa through ethical research protocols, as it is mandatory to concur ethical dimensions due to the sensitive nature of VMHAs. The lack of \textit{ethical dimensions} in WoeBot and Wysa is exemplified through non-clinical grounding and lack of contextual awareness in responses to the emergencies such as disclosure of immediate harm or suicidal ideation~\cite{koutsouleris2022promise}. To this end, the development of \textit{safe and explainable} VMHAs shall enhance their capabilities of reading between the lines resulting in accountable and fair conversational agents. For a well-aware (about user's depression) dialogue agent, it is perhaps \textit{safer} to avoid mentioning or inquiring about the topics that can worsen the users' mental health condition~\cite{henderson2018ethical}. 

Although WoeBot employs medical and process knowledge, to \textbf{explain} the decision-making, we investigate the relevant datasets for FAIR principles\footnote{\url{https://www.go-fair.org/fair-principles/}} (see Table~\ref{tab:datasets}) and evaluation metrics for quantitative and qualitative performance analysis of VMHAs' question-response generation module in active communication~\cite{brocki2023deep}. We further investigate existing evaluation metrics from \textit{passive communication} to support the VMHAs for \textit{active communication}.

\section{Discussion}

The field of AI-powered automated VMHAs is still in its nascent phase and continuously evolving to provide accessible health care to an increasing number of patients with mental illnesses. However, repetitive question/answer functionality within the models fails to sustain the user's engagement. 
Irrespective of deploying state-of-the-art VMHAs to mitigate the problems of the overburdened healthcare systems, the gap still remains between \underline{user's clinical needs} and \underline{VMHAs} that is yet to be connected. Despite the significant amount of studies 
in realizing the requirement of \textit{safety, harmlessness, explainability, curation of process and medical knowledge-based datasets and knowledge-infused learning methods}, they have never been incorporated or evaluated to enhance the contextualized conversations within a VMHA and their role in the emerging areas of mental healthcare. Hence, there is an urgent need to incorporate high-level contextual analysis and infuse new technical abilities of AI for VMHA. We outline two sub-sections to discuss: (i) the need of revamping the \textit{evaluation metrics}, and (ii) \textit{emerging} areas for developing safe and explainable VMHAs. 

\subsection{ Evaluation Method} 
\label{eval}
All the notable earlier work \cite{walker1997paradise} included subjective measures involving human-in-the-loop to evaluate a conversational system for its utility in the general purpose domain. Due to the expensive nature of human-based evaluation procedures, researchers have started using machine learning-based automatic quantitative metrics (e.g., BLEURT, BERTScore~\cite{clinciu2021study}, BLEU \cite{papineni2002bleu}, ROUGE \cite{lin2004rouge}) to evaluate the semantic similarity of the machine-translated text. ~\cite{liu2017improved} highlights the disagreement of users with existing metrics thereby lowering their expectations. Also, most of these traditional quantitative metrics are reference-based which is limited in availability and very difficult to ensure the quality of the human-written references~\cite{bao2022docasref}. To address these issues and holistically evaluate a desired VMHA with respect to \textit{explainability}, \textit{safety}, and \textit{knowledge process inclusion}, we need to revamp the metrics to bring VMHA systems closer to real-time applications.


\paragraph{Qualitative Metric}  We define mutlimetric evaluation strategy by instilling metrics that correlate well with human judgement and can provide more granular analysis towards more realistic VMHAs.

\begin{itemize}
    \item \textbf{Adherence:}
    Adherence, a long-standing discussion in the healthcare sector~\cite{fadhil2018conversational}, is defined as a commitment towards the goal (e.g., long-term therapy, physical activity, or medicine). 
    Despite the AI community showing a significant interest in evaluating the adherence of users~\cite{davis2020process} towards health assistants, 
    the lack of \textit{safe} response, in terms of \textit{DiG} and \textit{UsEx} in VMHAs, add the criticism with loss of adherence. This situation necessitates the requirement of adherence as a qualitative metric towards realizing more \textit{realistic} and \textit{contextual} VMHAs while treating patients with serious mental illness. 
    
    \item \textbf{Harmlessness:} 
    The conversational agents tend to generate harmful, unsafe, and sometimes incoherent information~\cite {welbl2021challenges}. Although researchers have made many efforts to curb the toxicity in the proliferation of hateful speech and biases in social media, much need to be realized when VMHAs are trained using the same datsets. 
    
   \item \textbf{Transparency:}
   The transparency and interpretability for understandable models (TIFU) framework emphasize the “explainability” of VMHAs by focusing on \textit{UsEx} and \textit{DiG}, thereby processing the knowledge to obtain clinically-verified responses \cite{joyce2023explainable}. 
\end{itemize}

\paragraph{KI Metric:}
In this section, we provide metrics that describe \textit{DiG}, \textit{safety},  \textit{MK} and \textit{PK} in table \ref{tab:my_label}. \ding{51} and \ding{55} tell whether VHMA has been tested for these KI metrics. 

\begin{itemize}
    \item \textbf{Safety:}
    Even though the datasets have been verified to be safe~\cite{sezgin2022operationalizing}, it is quite difficult to evaluate the models based on acceptable standards of safety because of their black-box nature. To include safety as a metric of evaluating conversational models,~\cite{roy2022proknow} introduces a safety lexicon as a glossary of clinical terms that the MHP would understand in their dataset. 
   ~\cite{henderson2018ethical} emphasizes on the underlying bias of a data-driven model and the need of an idea for contextual safety in the dialogue systems. 
 
    \item \textbf{Logical Coherence (LC):} 
    LC is a qualitative check of the logical relationship between a user's input and the follow-up questions measuring \textit{PK} and \textit{MK}. 
   ~\cite{kane2020nubia} used LC to ensure the reliable output from the RoBERTa model trained on the MNLI challenge and natural language inference GLUE benchmark, hence, opening new research directions towards safer models for MedNLI dataset~\cite{romanov2018lessons}. 

    \item \textbf{Semantic Relations (SR):}
    SR measures the extent of similarity between the response generation and the user's query~\cite{kane2020nubia}. \cite{gaur2022iseeq} highlights the use of SR for logical ordering of question-generations and hence, preventing language models from hallucinations. 
    It further enhances the use of VMHAs for profiling the user's illness and generating appropriate responses through \textit{DiG}.

\end{itemize}

\subsection{Emerging Areas of VMHAs}

\paragraph{Mental Health Triage}

Mental Health Triage\footnote{\url{https://en.wikipedia.org/wiki/Mental_health_triage}} is a risk assessment 
that categorizes the severity of the mental disturbance before suggesting psychiatric help to the users and categorizes them on the basis of urgency.
The screening and triage system could fulfill more complex requirements to achieve automated triage empowered by AI. A recent surge in the use of screening mechanisms by Babylon\footnote{https://tinyurl.com/2p8be7d4} and Limbic\footnote{https://tinyurl.com/2s44uxnk} has given new research directions towards a \textit{trustworthy} and \textit{safe} models in near future.


\noindent \paragraph{Motivational Interviewing}


Motivational Interviewing (MI) is a directive, user-centered counseling style for eliciting behavior change by helping clients to explore and resolve ambivalence. In contrast to the assessment of severity in mental health triaging, MI enables more interpersonal relationships for cure with a possible extension of MI for mental illness domain~\cite{westra2011extending}.~\cite{wu2020towards} suggest human-like empathetic response generation in MI with support for \textit{UsEx} and \textit{contextualization} with clinical knowledge. Recent works in identifying the interpersonal risk factors~\cite{ghosh2022no} from offline text documents further support MI for active communications.






\paragraph{Clinical Diagnostic Interviewing (CDI)}: 
CDI is a direct client-centered interview 
 between a clinician and patient without any
intervention. With multiple modalities of the CDI data (e.g., video, text, audio), the applications are developed in accordance with Diagnostic and Statistical Manual of Mental Disorders (DSM-V) to facilitate a quick gathering of detailed information about the patient. In contrast to the in-person sessions (leveraged on both verbal and non-verbal communication), the conversational agents miss the \textit{personalized} and \textit{contextual} information from non-verbal communication hindering the efficacy of VMHAs.   
 


\subsection{Practical Considerations}

We now consider two  practical considerations with VMHAs.

\noindent {\bf Difference in human v/s machine assistance:} For the VMHAs to be accepted by people in need, it is important that they feel the output of the system is valuable and useful. If the user had sought the help of a human mental health professional in the past, she would expect a similarly realistic conversational experience from the VMHA. However, getting training data from real conversations is expensive and fraught with data privacy and annotation challenges. To maintain the confidentiality of user data,  approaches akin to popular methods used in recommendation literature for creating training data from user data could be used: (a)  anonymize real data, (b) abstract from real data to create representative (but inaccurate)  samples, and (c) generate synthetic conversations based on characteristics of real data. In recommendations, user data is used to create personas while in the case of VMHA, real conversations can be used to create conversation templates and assign user profiles~\cite{user-profile-conversations}. But high-quality annotations on (conversation) data are a more significant problem and widespread in learning-based AI tasks.

\noindent {\bf Perception of quality with assistance offered:} A well-understood result in marketing is that people perceive the quality of a service based on the price paid for it as well as the word of mouth buzz around it~\cite{service-quality-price}. In the case of VMHAs, it is an open question whether help offered by VMHAs will be considered inferior to that offered by professionals. More crucially, if a user perceives it negatively, will this further aggravate the user's mental condition?

\section{Conclusion}
From 297 studies on mental health (active and passive communications), we present a systematic survey of $\sim$ 80 intelligent technologies for improving the user experience through VMHA or potential VMHA. 
We first propose a taxonomy of the mental healthcare domain for social NLP research, thrusting on benchmarking evaluation metrics for active communications. We then discussed the efforts in knowledge-driven AI for mental health, its connection with  \textit{UsEx} and \textit{safety}, and provided methods of improving and enhancing VMHA to support triaging, motivational interviewing, and diagnostic interviews. Finally, the survey sees its extension to ``personalization'' in VMHA, which is needed to perform tasks like screening, triaging and MI. Recently, Anthropic's Claude(a competitor of ChatGPT) is another effort to induce better safety with UsEx in the conversational system \cite{bai2022constitutional}.  

\section*{Ethical Statement} 
We adhere to anonymity, data privacy, intended use, and practical implication of the VMHAs. The questionnaires and rating scales described as clinical process knowledge do not contain personally identifiable information. The datasets covered in the survey are publicly available and can be obtained from user-author agreement forms. The text conversation in the figures are abstract and has no relevance with the real-time data source or any person.

\footnotesize
\bibliographystyle{named}
\bibliography{main.bib}

\end{document}